\documentclass[runningheads]{llncs}
\usepackage[T1]{fontenc}
\usepackage{graphicx}
\usepackage{amssymb}
\usepackage{algorithm}
\usepackage{algorithmicx}
\usepackage{algpseudocode}
\usepackage{tikz}
\usepackage[colorlinks,linkcolor=blue]{hyperref}
\usepackage{booktabs}
\usepackage{xcolor}
\usepackage{amsmath}
\usepackage{tabto}
\usepackage{bm}
\graphicspath{images}
\definecolor{jancolor}{rgb}{0.4,0.6,0.2}
\definecolor{hhcolor}{rgb}{0.2,0.6,0.6}
\definecolor{brcolor}{rgb}{0.9,0.1,0.1}
\definecolor{tscolor}{rgb}{1.0, 0.6, 0.2}
\definecolor{todocolor}{rgb}{0.9,0.1,0.1}
\definecolor{changedcolor}{rgb}{0.42,0.27,0.57}

{}
{}

\newcommand{\AlgoParams}[2]{
    \foreach\x/\y [count=\i] in {#1} {
            \Statex\ifnum\i=1 \textbf{#2} \fi \tabto*{1.6cm} \x \tabto*{4cm} \y
        }
}
\renewcommand{\Require}[1]{\AlgoParams{#1}{Require}}

\usepackage{graphicx,color}
\usepackage{atbegshi,picture}

\AtBeginShipoutNext{\AtBeginShipoutUpperLeft{%
\setlength{\unitlength}{1pt}%
\protect{\put(.5\paperwidth,-.5\paperheight){%
\makebox(-515,0){\rotatebox{90}{\textcolor[gray]{0.90}%
   {\Large Presented as workshop paper @ ECML PKDD 2024}}}%
}}}}

\begin{document}
\title{Edge-Based Graph Component Pooling}
%
\author{
    T. Snelleman    \inst{1} \and
    B.M. Renting    \inst{2,3} \and
    H.H. Hoos       \inst{1,2} \and
    J.N. van Rijn   \inst{2}
}

 \authorrunning{Snelleman et al.}
%
\institute{Chair of AI Methodology (AIM), RWTH Aachen University,\\ 52062 Aachen, Germany\\ \email{\{snelleman, hh\}@aim.rwth-aachen.de}\and
 Leiden Institute for Advanced Computer Science (LIACS), Leiden University,\\
 2333 CA Leiden, The Netherlands\\
 \email{\{b.m.renting, j.n.van.rijn\}@liacs.leidenuniv.nl}\and
 Delft University of Technology\\ 2628 XE Delft, The Netherlands
}
\maketitle              
%
\begin{abstract}

Graph-structured data naturally occurs in many research fields, such as chemistry and sociology. The relational information contained therein can be leveraged to statistically model graph properties through geometrical deep learning. Graph neural networks employ techniques, such as message-passing layers, to propagate local features through a graph. However, message-passing layers can be computationally expensive when dealing with large and sparse graphs. Graph pooling operators offer the possibility of removing or merging nodes in such graphs, thus lowering computational costs. However, pooling operators that remove nodes cause data loss, and pooling operators that merge nodes are often computationally expensive.

We propose a pooling operator that merges nodes so as not to cause data loss but is also conceptually simple and computationally inexpensive. We empirically demonstrate that the proposed pooling operator performs statistically significantly better than edge pool on four popular benchmark datasets while reducing time complexity and the number of trainable parameters by $70.6\%$ on average. Compared to another maximally powerful method named Graph Isomporhic Network, we show that we outperform them on two popular benchmark datasets while reducing the number of learnable parameters on average by $60.9\%$.

\end{abstract}

\section{Introduction}
\label{sec:introduction}
Graph-structured data is commonly encountered in many fields of study~\cite{bronstein2017geometric}, such as biology, chemistry, and sociology. Over the past few years, graph neural networks have seen a multitude of new graph-based operators to improve graph or node classification tasks. The operators can generally be categorised as message-passing, such as graph convolutional networks~\cite{kipf2017semi}, graph isomorphism networks~\cite{xu2018powerful} and GraphSAGE~\cite{hamilton2017inductive}, and pooling, such as node cluster pooling~\cite{diehl2019towards,khasahmadi2020memorybased,ying2018hierarchical,yuan2020structpool} and node drop pooling~\cite{gao2019graph,gao2021topology,lee2019self,zhang2020structure}.

Message-passing operators propagate the features of a node to its local neighbourhood with a distance of $k$ hops per operator, the exception being sequential message-passing operators specifically designed for directed acyclic graphs~\cite{thost2021directed}. They directly update node features by a permutation-invariant aggregation function, as the neighbourhoods of nodes in a graph are generally unordered. Although effective, these methods can become costly for large and sparse graphs, as many iterations of an operator may be required for information to propagate over the longest path in the graph.

Graph pooling operators aim to improve computational performance by modifying the structure of the given graph by removing or combining a subset of its nodes. In graph neural networks, these operators are used to, e.g., create more adequate graph-level representations~\cite{liu2023graph}. Pooling operators can offer several advantages when working with deep learning on graphs, such as improving computational efficiency by reducing the number of nodes and edges or by making the graph easier to interpret. They can also improve the connectivity in the graph by reducing the longest path so that messages are propagated further in a single layer. This can be crucial for performance when handling large, sparsely connected graphs~\cite{bianchi2024expressive,liu2023graph}. Liu et al.~\cite{liu2023graph} established a taxonomy of graph pooling operators. They defined two subcategories of graph pooling: Node cluster pooling and node drop pooling; the key difference between these is that the latter sacrifices information in the network for speed. In this work, we aim to bridge this gap and create a relatively efficient graph pooling operator that does not drop nodes and thus reduces information loss.

We propose an improved version of edge contraction pooling by Diehl et al.~\cite{diehl2019towards}, named \emph{edge-based graph component pooling}, addressing several design limitations. Our main contributions are:

\begin{itemize}
    \item We remove the hard constraints of edge contraction pooling where always half the nodes are merged, and nodes can only be merged with one neighbour, creating a more flexible operator.
    \item Our new operator is computationally cheap, with worst-case time complexity quadratic in the number of nodes. 
    \item We show that our operator improves performance compared to edge contraction pooling while being substantially more computationally efficient.
    \item We show that our operator does not suffer information loss by obtaining comparable performance to an expensive graph neural network that does not pool nodes.
\end{itemize}

Our proposed method, graph component pooling, like edge contraction pooling, offers other advantages over node drop pooling methods. It maintains information on how nodes were combined, preserving their information and allowing the pooling step to be reverted. This is an essential characteristic for the layer to be relevant for node-based tasks. In contrast, node drop pooling methods cause inevitable loss of information and cannot be used in node-based tasks, but are generally more computationally efficient~\cite{liu2023graph}. Our method aims to provide the full benefits of node cluster pooling methods whilst approaching the efficiency of node drop pooling methods.

\section{Related Work}
Liu et al.~\cite{liu2023graph} introduce a generalized framework for pooling operators in graph neural networks. The authors note the necessity of graph pooling operators for graph-level representations and summarise the latest developments in the field~\cite{bianchi2024expressive,khasahmadi2020memorybased,yuan2020structpool,gao2021topology,zhang2020structure,itoh2022multi,kayxuan2023distribution}. They separate the process of graph pooling into either two (node cluster pooling) or three (node drop pooling) steps. Node cluster pooling consists of a cluster assignment matrix step, where the cluster assignment for each node is determined, and a graph coarsening step where the new graph is created based on the node features $X$ and the adjacency matrix $A$. Node drop pooling is constructed by a score generator step, followed by a node selector and finalized with a graph coarsening step based on the feature matrix $X$, the adjacency matrix $A$, the scores $S$ and the selected nodes $idx$. Liu et al.~\cite{liu2023graph} state that edge contraction pooling belongs to neither of these categories as it approaches the problem from the edge view, but we will go on to show that our method belongs to the node clustering pooling subgroup due to our adaptations. Liu et al.~\cite{liu2023graph} establish well-known datasets used for benchmarking graph neural networks, including the datasets used in our experiments. They also presented challenges and opportunities in the field, and in our work, we contribute to one of these: We present our operator with an unpooling component to support node-based tasks in the future, as Liu et al.~\cite{liu2023graph} quantifiably shows that currently most pooling approaches have been focussed on graph-based tasks rather than node-based tasks. Secondly, in terms of expressive power, we contribute a method that is maximally expressive, as Bianchi et al.~\cite{bianchi2024expressive} have shown that the work of Diehl et al.~\cite{diehl2019towards} is theoretically as expressive as the Weisfeller-Lehman (WL) test. Since we use the same aggregation method as the work of Diehl et al.~\cite{diehl2019towards}, this proof also holds up for our approach.

In geometric deep learning, Xu et al.~\cite{xu2018powerful} have outlined the mathematical capabilities and limitations of graph neural networks (GNNs). More importantly, in their work, they have shown that GNNs are at best as powerful as the WL graph isomorphism test~\cite{leman1968reduction}, an algorithm designed to test whether two graphs $G$ and $H$ are \emph{not} isomorphic. The strong relation between the WL test and GNNs lies in the parallelism that both are message-passing constructs: They iteratively create a representation of a node based on its local neighbourhood. To distinguish different graph structures, the combined representation of the nodes must be distinguishable if the graphs are not isomorphic~\cite{xu2018powerful}. Their proposed method, graph isomorphism network, is a message-passing layer constructed of a multi-layer perceptron to model a function to update a node's representation based on its local neighbourhood. The authors prove that their method is as powerful as the WL test.

Although our method does not fall within the message-passing framework, we designed it to align with the mathematical proofs of maximizing the expressiveness of the GNNs. Bianchi et al.~\cite{bianchi2024expressive} stated that a pooling operator can be considered as powerful as the message-passing layers before it, as long as the network remains as expressive as after the pooling operation. The original edge contraction pool~\cite{diehl2019towards} is considered to be maximally powerful, according to Bianchi et al.~\cite{bianchi2024expressive}, who also demonstrated its potential empirically. 
Specifically, they studied the expressiveness of pooling operators using a modified SAT dataset of graphs that are provably non-isomorphic by the WL test, and on this benchmark of pooling operators, edge contraction was the only pooling operator to achieve 100.0\% accuracy. 
Our work meets the theoretical criteria of Bianchi et al.\cite{bianchi2024expressive} by maintaining the injective property of the aggregation function for our clusters (supernodes). It is important to note that Bianchi et al.~\cite{bianchi2024expressive} also empirically observed high running times for edge contraction pool (in terms of wallclock time), an important challenge we aim to overcome in computational complexity, required number of layers and computational efficiency of the layer itself. We find that the theoretical and practical impact of the work of Xu et al.~\cite{xu2018powerful} has been of great significance in the field of geometric deep learning, and therefore, in \autoref{sec:experiments}, we compare the effectiveness of our methods on the benchmark datasets to the graph neural networks presented in Xu et al.~\cite{xu2018powerful}.

Our work is a direct extension of the method proposed by Diehl et al.~\cite{diehl2019towards}. They introduced edge contraction pooling, a technique that scores the edges in a graph and contracts those with the highest scores until either half of the nodes have been pooled together or no nodes are eligible anymore for pooling. This requires sorting the edge scores to determine the edge order, yielding an upper bound time complexity of $\mathcal{O}(n^2 \cdot \log(n))$ where $n$ represents the number of nodes, as the number of edges in a graph is at maximum the number of nodes squared. We found this technique to be overly restrictive. The two most important limitations are as follows: Edge pooling always attempts to reduce the number of nodes by 50\%. This is a relatively arbitrary fraction, as both the task and the network may require varying amounts of pooling based on the nature and context of the data. Secondly, the pooling operator is restricted in terms of nodes per cluster, namely no cluster may consist of more than two nodes. This can negatively impact the efficacy of the operator, as an edge with a relatively high score may be ignored in favour of an edge with a lower score due to one of its nodes already participating in a cluster. We address both of these limitations by introducing a more flexible pooling operator that selects the edges which exceed some user-configurable threshold. As a downside, we are not guaranteed to which extent, if at all, the size of the graph is reduced.  We are not the first to seek improvements in the work of Diehl et al.~\cite{diehl2019towards}: Landolfi~\cite{landolfi2022revisiting} has suggested parallelization of the edge contraction pooling method, an implementation improvement we have also integrated into our work.

\section{Methodology}
\label{sec:methodology}
In this paper, we will combine the notations from Bronstein et al.~\cite{bronstein2021book} and Liu et al.~\cite{liu2023graph}. 
Let $\mathcal{G} = \langle \mathcal{V}, \mathcal{E} \rangle$ be a graph with nodes $\mathcal{V} = \langle 1, \cdots, m \rangle$ and edges $\mathcal{E} =  \langle 1, \cdots, n \rangle$. 
Nodes have features $x_i$ with dimension $d$, such that the feature matrix is $X \in \mathbb{R}^{m \times d}$. Each edge $e \in \mathcal{E}$ is a pair of nodes $e = \langle i, j \rangle$, and we consider  the adjacency matrix $A \in \{0,1\}^{m \times m}$ with $A_{ij} = 1$ if $\langle i, j \rangle \in \mathcal{E}$ and $0$ otherwise. 
A pooling operator $\text{POOL}$ acts on a graph such that $\mathcal{G}' = \mathrm{POOL}(\mathcal{G})$, with $\mathcal{G}' = \langle \mathcal{V}',\mathcal{E}' \rangle$ where $|\mathcal{V}'| \leq |\mathcal{V}|$ and $|\mathcal{E}'| \leq |\mathcal{E}|$. 

\begin{figure}
    \centering
    \includegraphics[width=\textwidth]{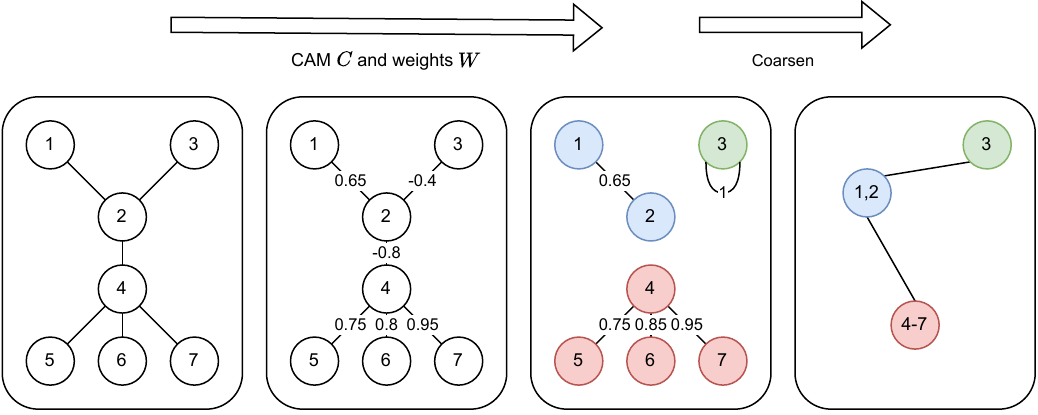}
    \caption{General illustration of our pooling operator $\mathcal{G}' = \mathrm{POOL}(\mathcal{G})$. Here we show how we create $\mathcal{G'} = \langle \mathcal{V'}, \mathcal{E'} \rangle$ from $\mathcal{G} = \langle \mathcal{V}, \mathcal{E} \rangle$ in two steps: We calculate the weights $W$ for every node using the edge scores and create the CAM $C$ using the graph component detection algorithm from Tarjan~\cite{tarjan1972depth}. We then coarsen the graph by combining the components into supernodes and removing the selected edges.
    \label{fig:method}}
\end{figure}

Lui et al.~\cite{liu2023graph} defined a two-step approach that can be identified in most graph pooling operators. The first step is to generate the cluster assignment matrix (CAM) $C$ that defines which nodes will be combined through binary or weighted values in the matrix. $C$ is often used as a multiplication factor for the original feature and adjacency matrices to transform the graph. The second step is graph coarsening, where the feature matrix $X$, the adjacency matrix $A$, and the CAM $C$ are used to obtain the new graph. Precisely how the CAM is obtained and coarsening is performed differs between graph pool operators. We will discuss our implementation of these two steps in the following sections. In \autoref{fig:method}, we summarize the overall structure of our pooling operator in a graphical representation.

\subsection{Cluster Assignment Matrix generator}\label{sec:methods:cam}
Our CAM is generated in three steps. The first step is defined in \autoref{eq:edgescore}. The concatenated features of two nodes connected by an edge are passed through a linear layer $\psi$, and a bias term $b$ is added. After applying an activation function $\sigma$, we obtain a score $S_{ij}$ used to decide whether two nodes should be merged. We treat undirected graphs as if both $\langle i, j \rangle \in \mathcal{E}$ and $\langle j, i \rangle \in \mathcal{E}$.
We acquire the subset of edges to be merged $\mathcal{E}_m$ using \autoref{eq:edgemerge} by setting a threshold $t$ for the edge score.

\begin{align}
    \label{eq:edgescore}
    S_{ij} &= \sigma \left( \psi(x_i, x_j) + b \right) \\ 
    \label{eq:edgemerge}
    \mathcal{E}_m &= \{ \langle i, j \rangle \ |\ S_{ij} > t \quad \forall \langle i, j \rangle \in \mathcal{E} \}
\end{align}

Clusters of nodes that will be merged are extracted by detecting weakly connected graph components~\cite{hopcroft1973efficient} based on $\mathcal{E}_m$. We use an algorithm with time complexity $\mathcal{O}(|\mathcal{V}|+|\mathcal{E}|)$, which can be reduced to $\mathcal{O}(|\mathcal{V}|^2)$ as $|\mathcal{E}|$ is upper bounded by $|\mathcal{V}|^2$. This algorithm was introduced by Tarjan~\cite{tarjan1972depth} and is implemented in Scipy~\cite{2020SciPy-NMeth}. The algorithm assigns each node to a cluster $\mathcal{V}_c$, resulting in a tuple of clusters $\langle \mathcal{V}_c^1, \cdots, \mathcal{V}_c^k \rangle$, such that $\mathcal{V} = \bigcup_{i = 1}^k \mathcal{V}_c^i$. Note that single nodes that are not merged are considered to be a cluster.
The cluster assignment matrix $C$ is created based on the detected clusters, where $C \in \{0,1\}^{|\mathcal{V}| \times k}$.

Separate from the cluster assignment matrix, we also create a weight matrix $W \in \mathbb{R}^{|\mathcal{V}| \times |\mathcal{V}|}$ that contains weights used to sum the features of nodes in a cluster. This also ensures that gradients from the loss function propagate backwards to the linear layer and bias term in \autoref{eq:edgescore}. All elements of $W$ are obtained by \autoref{eq:weightsmat}.

\begin{equation}\label{eq:weightsmat}
    W_{ij} = 
    \begin{cases}
        S_{ij} & \text{if } \langle i, j \rangle \in \mathcal{E}_m \\
        1 & \text{if } i = j \text{ and } v_i \text{ will not be merged} \\
        0 & \text{otherwise}
    \end{cases}
\end{equation}

\subsection{Graph Coarsening}
\label{sec:methods:gc}
Using the cluster assignment matrix $C$ and the weight matrix $W$, we can coarsen the graph by merging the clusters of nodes. Weighted sums are applied to combine the features of the merged nodes to prevent information loss. Starting with node feature matrix $X$ and adjacency matrix $A$, we obtain coarsened feature and adjacency matrices $X'$ and $A'$ through \autoref{eq:coarsen}.

\begin{align}\label{eq:coarsen}
\begin{split}
    X' &=  (W \times C)^\intercal \times X \\
    U &=  C^\intercal \times A \times C \\
    A' &= (\min(U_{ij}, 1))_{1 \leq i,j \leq k}
\end{split}
\end{align}

In \autoref{eq:coarsen}, the matrix $U$ can have values greater than 1, as it sums the number of edges between nodes that are merged into a cluster. Therefore, the $\min$ function is applied over the elements of matrix $U$ to obtain the updated adjacency matrix $A'$ containing only zeros and ones. Finally, we rebuild the set of nodes $\mathcal{V}'$ to match the coarsened feature matrix $X'$ and rebuild the set of edges based on the coarsened adjacency matrix $A'$ using \autoref{eq:edgesetcoarsen}.

\begin{equation}\label{eq:edgesetcoarsen}
    \mathcal{E}' = \{ \langle i, j \rangle \ |\ A'_{ij} = 1 \}
\end{equation}

The complete method is provided as pseudo-code in \autoref{alg:cap}. For details on the ``ConnectedComponents'' procedure, we refer the reader to Tarjan et al.~\cite{tarjan1972depth} and the documentation of its implementation in the SciPy package~\cite{2020SciPy-NMeth} used in this paper.\footnote{\url{https://docs.scipy.org/doc/scipy/reference/generated/scipy.sparse.csgraph.connected_components.html}}

\begin{algorithm}
    \caption{Edge-based Contraction Pooling algorithm}\label{alg:cap}
    \begin{algorithmic}[1]
    \Require{   {$X \in \mathbb{R}^{m \times d}$/Feature matrix},
                {$A \in \{0,1\}^{m \times m}$/Adjacency matrix},
                {$t \in  [0,1]$/Threshold}}
    
    \vspace{0.1cm}\hrule\vspace{0.1cm}

    \State $S \gets \bm{0}^{m \times m}$
    \For{\textbf{all} $i, j $ \textbf{such that} $A[i,j] = 1$}
        \State $S[i,j] \gets \sigma \left( \psi(x_i, x_j) + b \right)$ \Comment{$x_i, x_j \in X$}
    \EndFor
    \State $A_{merge} \gets S \geq t$ \Comment{1 where true, 0 otherwise}
    \State $W \gets S \circ A_{merge}$ \Comment{Hadamard product}
    \State $C \gets \text{ConnectedComponents}(A_{merge})$ \Comment{Tarjan et al.~\cite{tarjan1972depth}}
    \State $N_{unmerged} \gets (\sum_{i=1}^{m}A_{merge}[i,\cdot] + \sum_{j=1}^{m}A_{merge}[\cdot,j]) = 0$
    \State $W \gets W + Diagonal(N_{unmerged})$

    \State $X' \gets  (W \times C)^\intercal \times X$
    \State $U \gets  C^\intercal \times A \times C$
    \State $A' \gets (\min(U_{ij}, 1))_{1 \leq i,j \leq k}$
    \State{\Return $X', A'$}
\end{algorithmic}

\end{algorithm}

\section{Emperical Evaluation}
\label{sec:experiments}
To evaluate our new operator, we test it in a graph classification task on a multitude of datasets while closely following common experimentation practices in the field (e.g., Ying et al.~\cite{ying2018hierarchical} and Zhang et al.~\cite{zhang2020structure}).

\subsection{Datasets}
\label{sec:datasets}
We use several datasets collected as a benchmark by Morris et al.~\cite{morris2020tudataset}. In \autoref{tab:datasettable}, these datasets are listed alongside core characteristics. The protein dataset is a classification problem where the task is to predict whether a protein acts as an enzyme. There are six social network datasets. The first is Collaboration, a set of scientific collaboration networks where the task is to predict the subdomain of research of a researcher based on their network. In the Reddit datasets, the graphs are user interactions in a subreddit, and the task is to predict the subreddit. The IMDB datasets describe movie collaborations of actors within a given set of genres, and the task is to predict the genre of such a collaboration network. Finally, the molecule dataset NCI1 was built to predict whether a chemical compound can suppress the growth of tumours in the human body.

Except for the proteins dataset, the datasets do not contain node features; thus, distinguishing the classes must be done based on the structure of the given graph. To allow for comparison with Xu et al.~\cite{xu2018powerful}, we have synthesized node features of the IMDB datasets, Collaboration and NCI1 by using a one-hot encoding of the nodes' neighbourhood size. For all other graphs, we have set the node features to a scalar of one, following Xu et al~\cite{xu2018powerful}. While Xu et al.~\cite{xu2018powerful} presented results on several other benchmark datasets, such as MUTAG and PTC, we decided not to include these datasets, as they are small and tend to yield unreliable results.

\begin{table}
    \centering
    \begin{tabular}{lcrrrcr}
        \toprule
        Dataset & Category & Size & Avg. $|\mathcal{V}|$ & Avg. $|\mathcal{E}|$ & Features & \# Classes\\
        \midrule
        Proteins~\cite{dobson2003distinguishing} & Protein & 1,113 & 39.06 & 72.82 & 8 & 2\\
        Reddit-Binary~\cite{yanardag2015deep} & Social & 2,000 & 429.63 & 497.75 & Scalar: 1 & 2\\
        Reddit-Multi-12K~\cite{yanardag2015deep} & Social & 11,929 & 391.41 & 456.89 & Scalar: 1 & 11\\
        Collaboration~\cite{yanardag2015deep} & Social & 5,000 & 74.49 & 2457.78 & Synthesized & 3\\
        IMDB Binary~\cite{yanardag2015deep} & Social & 1,000 & 19.77 & 96.53 & Synthesized & 2\\
        IMDB Multi~\cite{yanardag2015deep} & Social & 1,500 & 13.00 & 65.94 & Synthesized & 3\\
        NCI1~\cite{wale2008comparison} & Molecule & 4,110 & 29.87 & 32.30 & Synthesized & 2\\
        Reddit-Multi-5K~\cite{yanardag2015deep} & Social & 4,999 & 508.52 & 594.87 & Scalar: 1 & 5\\
        \bottomrule
    \end{tabular}
    \caption{Dataset metadata, as described by Liu et al~\cite{liu2023graph};$|\mathcal{V}|$ denotes the average number of nodes in a graph, and $|\mathcal{E}|$ is the average number of edges. The features column describes node-level features. For graphs without node features, we set their features to scalar 1.}
    \label{tab:datasettable}
    \vspace{-3mm}
\end{table}

\subsection{Setup of Experiments}
Xu et al.~\cite{xu2018powerful} and Diehl et al.~\cite{diehl2019towards} trained their method 10 times on randomly split training and validation sets, using $10$-fold cross-validation. Xu et al.~\cite{xu2018powerful} report performance on the validation set, Diehl et al.~\cite{diehl2019towards} on the test set. However, the limited number of iterations can lead to unreliable performance measures as some datasets are small (e.g., the protein dataset), causing skewed label distributions in the sampled sets. We mitigate this by repeating this procedure 100 rather than 10 times. We re-evaluate our and their methods with the extended number of training iterations. The random seed used to split the sets is re-used in each experiment, ensuring that all methods use identical data.

We recreated the method of Xu et al.~\cite{xu2018powerful} based on the architectures and hyperparameters described in their paper. The pooling operator designed by Diehl et al.~\cite{diehl2019towards} is included in the PyTorch Geometric~\cite{fey2019fast} package, and we re-implemented the remainder of their architecture based on the paper using SageConv~\cite{hamilton2017inductive} and optimized the hyperparameters manually. We do not evaluate their methods on all datasets, as they did not design a network for all of the datasets. Designing such networks ourselves would lead to an unfair comparison.

Like Diehl et al.~\cite{diehl2019towards}, we loosely based the architectures of our models on those of Xu et al.~\cite{xu2018powerful} and Fey and Lenssen~\cite{fey2019fast}. For our pooling operator, we set $\sigma$ in \autoref{eq:edgescore} to be the hyperbolic tangent activation function and set $t = 0$ in \autoref{eq:edgemerge}. In most architectures, dropout is applied to mitigate overfitting on the training datasets~\cite{srivastava2014dropout}. To allow for graph-level classification, we apply global sum pooling on the node feature matrix after the graph convolutional layers to reduce it to a vector. Finally, the obtained vector is passed through linear layers to obtain the classification logits. We present the most important hyperparameters of our models in \autoref{tab:hptable}. Our code implementation can be found on Github.\footnote{\href{https://github.com/ADA-research/graphclusterpool}{https://github.com/ADA-research/graphclusterpool}}

\begin{table}
    \centering
    \resizebox{\textwidth}{!}{
    \begin{tabular}{llrrrrrr}
        \toprule
        Variable & Architecture & Hidden size & \# Epochs & LR & LR halving & Dropout & \# Params\\
        \midrule
        Proteins & $C P C L$ & 16 & 200 & 0.001 & 100 epochs & 0.100 & 802\\
        Reddit-Binary & $C C P C C P C L L$ & 128 & 200 & 0.001 & 50 epochs & 0 & 83\,459\\
        Reddit-Multi 12K & $C C P C C P C L L$ & 256 & 200 & 0.00025 & 55 epochs & 0.025 & 333\,325\\
        Collaboration & $C P C L$ & 32 & 100 & 0.001 & 65 epochs & 0.500 & 12\,996\\
        IMDB Binary & $C P C L$ & 32 & 100 & 0.0001 & 22 epochs & 0.100 & 18\,498\\
        IMDB Multi & $C P C L$ & 128 & 200 & 0.001 & 45 epochs & 0.100 & 62\,468\\
        NCI1 & $C P C L$ & 128 & 200 & 0.005 & 45 epochs & 0.100 & 38\,274\\
        Reddit-Multi 5K & $C C P C C P C L L$ & 128 & 300 & 0.0007 & 80 epochs & 0 & 83\,975\\
        \bottomrule
    \end{tabular}}
    \caption{Hyperparameters of the models. The architecture is described by letters from left to right, where $C$ is a graph convolutional layer~\cite{kipf2017semi} with ReLU activation function, $P$ is our pooling layer with hyperbolic tangent activation, and $L$ is a fully connected linear layer with ReLU activation. The output size of the intermediate layers is identical for every layer of the architectures, with the exception of the output layer, which is set to match the number of classes in the classification task considered. Here, the sigmoid activation function is used for binary classification and the softmax activation function for multi-class classification. }\label{tab:hptable}
\end{table}

We split the datasets into training, validation, and test sets of 80\%, 10\% and 10\% of the data, respectively. The validation set is used to make train-time decisions such as outputting the best model based on past validation scores. The reported performance is based on 100 evaluations on the test set. 

\subsection{Results}
\begin{figure}
    \centering
    \includegraphics[width=\textwidth]{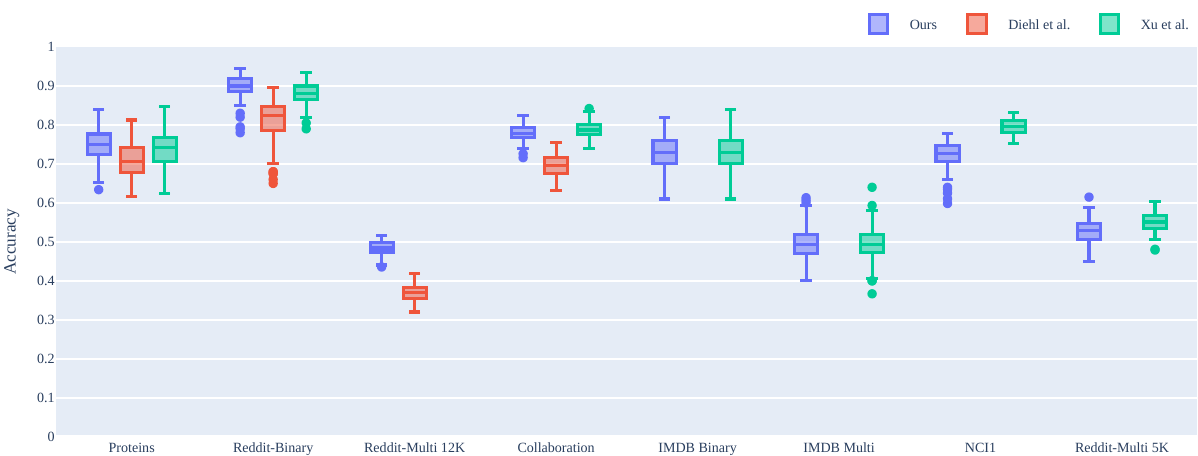}
    \caption{Boxplot visualization of the test set results shown in \autoref{tab:dataset_results}, with accuracy on the y-axis. The results of our method and the methods we compare against are grouped per benchmark dataset on the x-axis.
    \label{fig:boxplot_dataset_results}}
\end{figure}

We present our results on the benchmark datasets in \autoref{tab:dataset_results} and \autoref{fig:boxplot_dataset_results}. 

\begin{table}
    \centering
    \resizebox{\textwidth}{!}{
    \begin{tabular}{lcccccccc}
        \toprule
        Dataset & Proteins & Reddit-Binary & Reddit-Multi 12K & Collaboration & IMDB Binary & IMDB Multi & NCI1 & Reddit-Multi 5K\\
        \midrule
        Ours & $\mathbf{74.7 \pm 3.9}$  & $\mathbf{89.7 \pm 3.0}$ & $\mathbf{48.4 \pm 1.7}$ & $77.9 \pm 2.0$ & \underline{$72.7 \pm 3.9$} & \underline{$49.6 \pm 4.3$} & $72.2 \pm 3.5$ & $52.6 \pm 3.0$\\
        Diehl et al. & $70.9 \pm 4.6$ & $81.1 \pm 5.6$ & $36.9 \pm 2.1$ & $69.5 \pm 2.7$ & - & - & - & -\\
        Xu et al. & $73.5 \pm 4.6$ & $87.8 \pm 2.7$ & - & $\mathbf{78.7 \pm 2.0}$ & \underline{$72.7 \pm 4.3$} & \underline{$49.6 \pm 4.3$} & $\mathbf{79.5 \pm 2.0}$ & $\mathbf{55.1 \pm 2.4}$\\
        \bottomrule
    \end{tabular}}
    \caption{Results on benchmark datasets~\cite{morris2020tudataset}; we report means and standard deviations over 100 test scores. Missing results are caused by the model design missing from the original work. Statistically significant results are marked in \textbf{boldface}, and statistically insignificant results are \underline{underlined}.}\label{tab:dataset_results}
    \vspace{-6mm}
\end{table}

In \autoref{tab:dataset_results}, we see a substantial improvement using our method compared to the work of Diehl et al.~\cite{diehl2019towards} on every benchmark dataset while improving the pooling operator to quadratic time complexity. Comparing our method to Xu et al.~\cite{xu2018powerful}, we observe an accuracy score improvement for Proteins and the Reddit-Binary benchmark. However, we see a decrease in performance on the NCI1 and Reddit-Multi-5K datasets.

We assessed the statistical significance of performance differences using a two-tailed t-test with a standard significance level of $0.05$ based on the 100 test scores. We compare our method to that of Xu et al.~\cite{xu2018powerful} and Diehl et al.~\cite{diehl2019towards}, report the $p$-values from the t-test in \autoref{tab:dataset_pvalues}, and indicate them in \autoref{tab:dataset_results}. 

We find that our results are significantly better for two datasets in comparison to Xu et al.~\cite{xu2018powerful}. On Proteins and Reddit Binary, we show an improvement of $1.2\%$ and $1.9\%$ on average, respectively. Furthermore, we've found that our results are significantly worse on the Collaboration, NCI1, and Reddit Multi-Classification 5K dataset, we show a decrease of $0.8\%$ $7.3\%$ and $1.9\%$, respectively. We find our results to be tied in performance for the IMDB datasets.

\begin{figure}
    \centering
    \includegraphics[width=\textwidth]{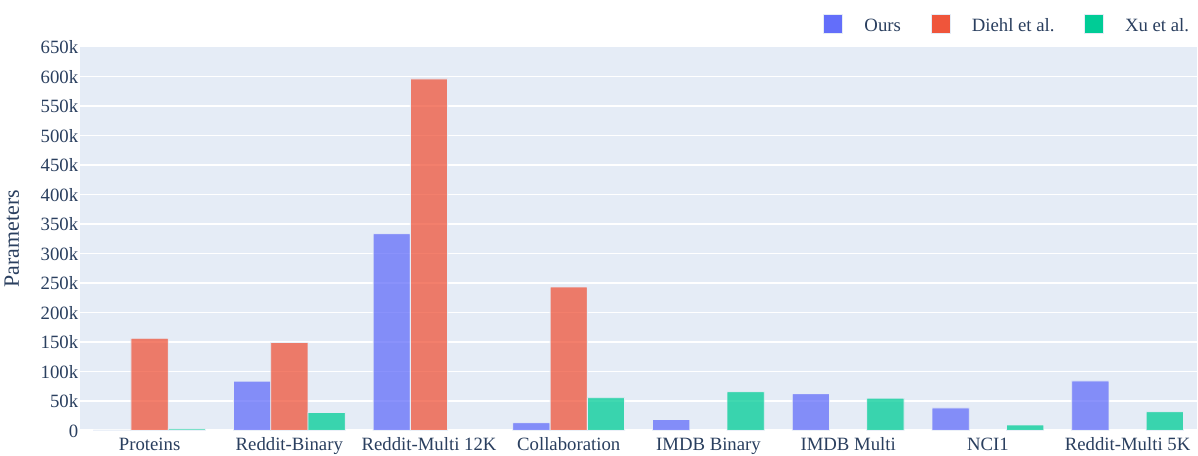}
    \caption{Bar graph of the number of learnable parameters used per model, per benchmark, as seen in \autoref{tab:dataset_results_params}. On the y-axis, we show the number of learnable parameters per model architecture of our method and the methods we compare against.
    \label{fig:barplot_dataset_results_params}}
\end{figure}

Finally, we show the number of parameters used per model in \autoref{fig:barplot_dataset_results_params}. 

We see a very substantial decrease of learnable parameters in comparison to Xu et al.~\cite{xu2018powerful} in three datasets, ranging from $70.8\%$ to $76.6\%$. However, in three cases a very substantial increase of learnable parameters can be found, averaging $159.2 \%$, showing that in this category our method struggles as well to compete with the method of Xu et al.~\cite{xu2018powerful}

The complete information on learnable parameters can be found in \autoref{tab:dataset_results_params}.

\begin{table}
    \centering
    \resizebox{\textwidth}{!}{
    \begin{tabular}{lrrrrrrrr}
        \toprule
        Dataset & Proteins & Reddit-Binary & Reddit-Multi 12K & Collaboration & IMDB Binary & IMDB Multi & NCI1 & Reddit-Multi 5K\\
        \midrule
        Diehl et al. & $0.000$ & $0.000$ & $0.000$ & $0.000$ & - & - & - & -\\
        Xu et al. & $0.048$ & $0.000$ & - & $0.009$ & $0.892$ & $0.887$ & $0.000$ & $0.000$\\
        \bottomrule
    \end{tabular}}
    \caption{$p$-values from the statistical significance test on performance differences between our results on benchmark datasets from \autoref{tab:dataset_results} and the works of Diehl et al.~\cite{diehl2019towards} and Xu et al~\cite{xu2018powerful}. We used a two-tailed t-test over a population of 100 test scores and rounded to three decimals. We reject the null hypothesis of equal performance if $p < 0.05$. We find that all our results are significantly different from Diehl et al.~\cite{diehl2019towards} and for Proteins, Reddit-Binary, Collaboration, NCI1 and Reddit-Multi 5K compared to Xu et al.~\cite{xu2018powerful}
    \label{tab:dataset_pvalues}}
    \vspace{-5mm}
\end{table}

\section{Conclusions \& Future Work}
In this work, we introduced a computationally efficient graph pooling operator, dubbed edge-based graph component pooling, based on the work of Diehl et al.~\cite{diehl2019towards}. We showed that the operator, like the original work, is a maximally expressive pooling layer~\cite{bianchi2024expressive}, thus, when added into any existing graph neural network, the expressiveness of the model is not reduced. When evaluating the performance of our operator, we showed that it achieves substantially better accuracy than that of Diehl et al.~\cite{diehl2019towards}.

We also compare to a method that does not apply pooling, the work of Xu et al.~\cite{xu2018powerful}, to see whether our pooling operator causes information loss. We showed that our method achieves statistically significant improved performance on two out of seven benchmarks and is significantly outperformed on three. On two other datasets, the performance of the models is similar. The comparable performance indicates that our operator does not cause information loss while having the benefit of reducing the number of required parameters through graph coarsening. Reducing the number of learnable parameters in a model has several advantages, such as mitigating overfitting~\cite{srivastava2014dropout}, reducing the computational cost for training and inference procedures, and avoiding using large storage space. Specifically, we have shown that, on average, our models use $70.6\%$ fewer learnable parameters compared to the original method of Diehl et al.~\cite{diehl2019towards}, highlighting the efficacy of our pooling operator.

Based on the work of Diehl et al.~\cite{diehl2019towards} and the presented opportunities in Liu et al.~\cite{liu2023graph}, we also implemented a reverse operator. Given the reduced graph, all the removed nodes and their respective edges can be restored by copying the cluster features into each node. This also allows the operator to be used for node classification tasks. It would be interesting to see how the operator would perform in, for example, an adapted graph U-Net structure~\cite{gao2019graph} for node-based tasks described by Liu et al.~\cite{liu2023graph}. We leave this for future work.

We did not consider edge features in this work, but they can trivially be included in our method. An approach for this would be to score the edges not only based on the node features but also include the edge features. This allows the edge features to have a direct impact on the features of the newly created node.

Our method uses a learned edge scoring method with a customisable threshold determining which edges will be merged, leading to an unknown number of edges being merged. This could be misaligned with the user's objective and reason to use edge pool~\cite{diehl2019towards} or top-k pool~\cite{gao2019graph} instead. An adaptation could be made to merge only part of the edges based on the edge scores, such as the highest quantile. Although interesting, we did not see this within the scope of our work, as we aimed to move away from hard proportionality reduction towards an unrestricted number of merged edges.

Finally, instead of pooling entire graphs by averaging all node features, which is sometimes done in deep learning on graphs, it would be interesting to see if our operator could pool entire graphs through the learned edge weights. This creates a learnable global pool operator. Whether such a learned global pool operator would perform well remains an open question.

%
%
%
\bibliographystyle{splncs04}
\bibliography{bibliography}
\newpage
\section*{Appendix}

\begin{table}[h]
    \centering
    \resizebox{\textwidth}{!}{
    \begin{tabular}{lrrrrrrrr}
        \toprule
        Dataset & Proteins & Reddit-Binary & Reddit-Multi 12K & Collaboration & IMDB Binary & IMDB Multi & NCI1 & Reddit-Multi 5K\\
        \midrule
         Diehl et al. & 156\,291 & 149\,123 & 595\,725 & 243\,077 & - & - & - & - \\
         Xu et al. & 2742 & 30\,538 & - & 55\,584 & 65\,638 & 54\,646 & 9294 & 31\,586\\
         Ours & 802 & 83\,459 & 333\,325 & 12\,996 & 18\,498 & 62\,468 & 38\,274 & 83\,975\\
         Diehl \% change & $-99.5\%$ & $-44.0\%$ & $-44.1\%$ & $-94.7\%$ & - & - & - & -\\
         Xu \% change & $-70.8\%$ & $+173.3\%$ & - & $-76.6\%$ & $-71.8\%$ & $+14.3\%$ & $+311.8\%$ & $+165.9\%$\\
        \bottomrule
    \end{tabular}}
    \caption{Number of learnable parameters in each neural network architecture, per benchmark dataset~\cite{morris2020tudataset} of our models versus Xu et al~\cite{xu2018powerful}.
    \label{tab:dataset_results_params}}
\end{table}

\end{document}